%% file: main.tex
\documentclass{article}
\usepackage[T1]{fontenc}
\usepackage[utf8]{inputenc}

\usepackage{microtype}
\usepackage{graphicx}
\usepackage{subcaption}
\usepackage{booktabs}

\usepackage[hyphens]{url}
\usepackage{hyperref}

\usepackage[preprint]{icml2026}

\usepackage{amsmath}
\usepackage{amssymb}
\usepackage{mathtools}
\usepackage{amsthm}
\usepackage[capitalize,noabbrev]{cleveref}

\theoremstyle{plain}

\theoremstyle{definition}

\theoremstyle{remark}

\usepackage[textsize=tiny,disable]{todonotes}
\icmltitlerunning{Can Theoretical Physics Research Benefit from Language Agents?}

\usepackage{tcolorbox}
\usepackage[compat=1.1.0]{tikz-feynman} 

\usepackage{soul}
\definecolor{DIFadd}{RGB}{0,127,0}
\definecolor{DIFdel}{RGB}{220,20,60}

\usepackage{listings}
\usepackage{tikz}
\usetikzlibrary{calc,chains,positioning,backgrounds}
\usetikzlibrary{arrows.meta, positioning, shapes, chains, backgrounds, calc}
\tikzset{ten/.style={fill=tensorblue}}
\tikzset{tenred/.style={fill=tensorred}}
\tikzset{tengreen/.style={fill=green!50!black!50}}
\tikzset{tenpurp/.style={fill=tensorpurp}}
\tikzset{tengrey/.style={fill=black!20}}
\tikzset{tenorange/.style={fill=orange!30}}
\tikzset{u/.style={fill=blue!20,draw=black}}
\tikzset{w/.style={fill=green!50!black!80,draw=black}}
\definecolor{tensorblue}{rgb}{0.8,0.8,1}
\definecolor{tensorred}{rgb}{1,0.5,0.5}
\definecolor{tensorpurp}{rgb}{1,0.5,1}

\newcommand{\diagram}[1]{ \begin{array}{cc}\begin{tikzpicture}[scale=.5,every node/.style={sloped,allow upside down},baseline={([yshift=+0ex]current bounding box.center)}] #1 \end{tikzpicture} \end{array} }

\begin{document}

\twocolumn[
  \icmltitle{Can Theoretical Physics Research Benefit from Language Agents?}

  \icmlsetsymbol{equal}{*}
  \icmlsetsymbol{eqsup}{\textsuperscript{\dag}}

  \begin{icmlauthorlist}
    \icmlauthor{Sirui Lu}{mpq,mcqst}
    \icmlauthor{Zhijing Jin}{mpi}
    \icmlauthor{Terry Jingchen Zhang}{eth}
    \icmlauthor{Pavel Kos}{mpq,mcqst}
    \icmlauthor{J. Ignacio Cirac}{eqsup,mpq,mcqst}
    \icmlauthor{Bernhard Sch\"olkopf}{eqsup,mpi,eth}
  \end{icmlauthorlist}

  \icmlaffiliation{mpq}{Max-Planck-Institut f\"ur Quantenoptik, Garching, Germany}
  \icmlaffiliation{mcqst}{Munich Center for Quantum Science and Technology, M\"unchen, Germany}
  \icmlaffiliation{mpi}{Max-Planck-Institut f\"ur Intelligente Systeme, T\"ubingen, Germany}
  \icmlaffiliation{eth}{ETH Z\"urich, Z\"urich, Switzerland}
  \icmlcorrespondingauthor{Sirui Lu}{sirui.lu@mpq.mpg.de}

  \vskip 0.3in
]

\printAffiliationsAndNotice{$^\dag$ Equal Supervision}

\input{Sections/MainText}

\section*{Acknowledgments}
S.L. thanks Tao Shi, Xingyan Chen, and Xiao-Liang Qi for helpful discussion.
This work is partially supported by the T\"ubingen AI Center
and by the Machine Learning Cluster of Excellence, EXC number 2064/1 - Project number 390727645.

\newpage
\appendix
\onecolumn
\end{document}

%% file: Sections/MainText.tex
\input{Sections/1-intro}

\bibliographystyle{icml2026}

\input{main.bbl}
\input{Sections/appendix}

%% file: Sections/1-intro.tex
\begin{abstract}
Large Language Models (LLMs) are rapidly advancing across diverse domains, yet their application in theoretical physics remains inadequate. While current models show competence in mathematical reasoning and code generation, we identify critical gaps in physical intuition, constraint satisfaction, and reliable reasoning that cannot be addressed through prompting alone. Physics demands approximation judgment, symmetry exploitation, and physical grounding that require AI agents specifically trained on physics reasoning patterns and equipped with physics-aware verification tools.
We argue that LLM would require such domain-specialized training and tooling to be useful in real-world for physics research.
We envision physics-specialized AI agents that seamlessly handle multimodal data, propose physically consistent hypotheses, and autonomously verify theoretical results. Realizing this vision requires developing physics-specific training datasets, reward signals that capture physical reasoning quality, and verification frameworks encoding fundamental principles. We call for collaborative efforts between physics and AI communities to build the specialized infrastructure necessary for AI-driven scientific discovery.
\end{abstract}

\section{Introduction}
\label{sec:introduction}
Large Language Models (LLMs) represent a major advance at the forefront of artificial intelligence (AI), exhibiting remarkable proficiency in understanding natural language and performing increasingly complex reasoning tasks~\cite{Brown2020Language,Radford2018Improving,OpenAI2023GPT4,Team2024Gemini,Team2024Geminia,Anthropic2024Claude}. While impacting various sectors, their potential in fundamental scientific research is primarily exploratory~\cite{Adesso2023Ultimate,Binz2025How}. Physics, with its blend of abstract theoretical derivation, rigorous experimentation \& computation, observation of real-world phenomena and reliance on physical intuition, presents both unique challenges and fertile ground for LLM applications.

\textbf{Position: We argue that LLM agents, when appropriately adapted and integrated with domain-specialized training and tooling, could potentially serve as a promising technology for accelerating discovery in theoretical physics, with broader implications for computational and applied physics, provided their current limitations in rigorous reasoning, physical grounding, and reliability are systematically addressed through targeted interdisciplinary research.}

This position challenges the current paradigm where LLMs serve primarily as assistants for information retrieval and aggregation. We contend that LLMs may evolve into autonomous collaborators for physicists, augmenting capabilities from literature review and conceptual exploration, to computational simulation and data interpretation. However, realizing this potential requires concentrated effort informed by both frontline physicists and AI developers.
Supporting this cautious optimism is recent progress in LLM architecture, scale, and particularly advances in reasoning models~\cite{OpenAI2023GPT4,Anthropic2024Claude,Team2024Gemini,Team2024Geminia,DeepSeek-AI2025DeepSeekV3,Guo2025DeepSeekR1,Team2025Kimi} that demonstrate growing agency in multi-step problem-solving needed for AI-driven physics research.

\textbf{Overview of this work}
This paper is structured as follows.
\Cref{sec:taxonomy} introduces the taxonomy used in this paper, outlining a typical physics research workflow with an overview of the subtasks LLMs might assist with. \Cref{sec:skill_analysis_physics} provides an in-depth analysis of LLM capabilities for physics reasoning, categorized into mathematical skills, physics-specific reasoning beyond mathematics, code generation \& execution, and general research skills. \Cref{sec:llm_engineering_skills} discusses common LLM engineering techniques relevant to physics applications. \Cref{sec:future_llms} explores open directions and desirable future capabilities for next-generation LLM-powered systems to better assist physics research. Finally, \Cref{sec:conclusion} summarizes our position and offers a concluding perspective.
Sections on Risk, Limitations and Ethical Considerations are included in the appendix.
\section{Physics Research: An Overview}
\label{sec:taxonomy}
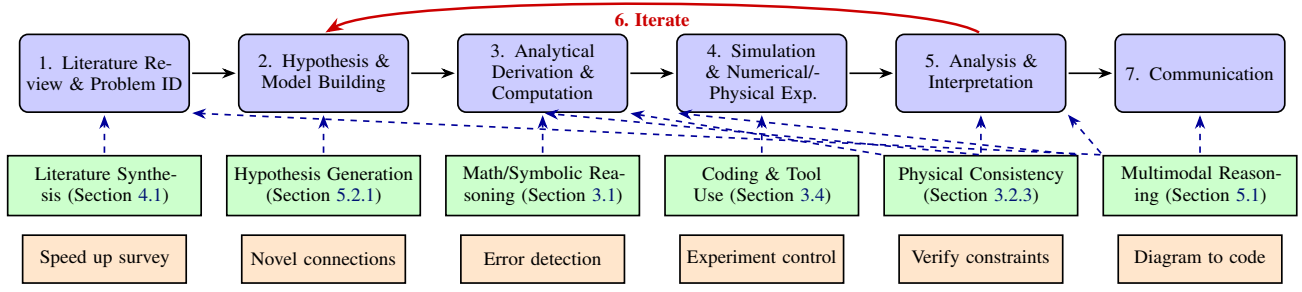
\begin{figure*}[t!]
    \centering
    \resizebox{\linewidth}{!}{%
        \begin{tikzpicture}[
                scale=0.7,
                node distance=0.2cm and 0.8cm,
                stage/.style={rectangle, rounded corners, draw=black, fill=blue!20, thick,
                        text centered, minimum height=1.3cm, text width=2.6cm, align=center, font=\small},
                skill/.style={rectangle, draw=black, fill=green!20, thick,
                        text centered, minimum height=1.0cm, text width=3.0cm, align=center, font=\small},
                opportunity/.style={rectangle, draw=black, fill=orange!20, thick,
                        text centered, minimum height=0.8cm, text width=2.5cm, align=center, font=\small},
                arrow/.style={-Stealth, thick, shorten >=1pt, shorten <=1pt},
                skillarrow/.style={-Stealth, thick, dashed, blue!60!black, shorten >=1pt, shorten <=1pt},
                iterationarrow/.style={-Stealth, thick, red!80!black, line width=1.2pt, shorten >=1pt, shorten <=1pt}
            ]
            \begin{scope}[start chain=workflow going right, node distance=0.8cm]
                \node (lit) [stage, on chain] {1. Literature Review \& Problem ID};
                \node (hyp) [stage, on chain] {2. Hypothesis \& Model Building};
                \node (der) [stage, on chain] {3. Analytical Derivation \& Computation};
                \node (sim) [stage, on chain] {4. Simulation \& Numerical/Physical Exp.};
                \node (ana) [stage, on chain] {5. Analysis \& Interpretation};
                \node (com) [stage, on chain] {7. Communication};
            \end{scope}
            \draw [arrow] (lit) -- (hyp);
            \draw [arrow] (hyp) -- (der);
            \draw [arrow] (der) -- (sim);
            \draw [arrow] (sim) -- (ana);
            \draw [arrow] (ana) -- (com);
            \draw [iterationarrow] (ana.north) to[out=150, in=30, looseness=0.3]
            node[below, midway, font=\small\bfseries, text=red!80!black] {6. Iterate} (hyp.north);
            \node (sk_lit) [skill, below=0.7cm of lit] {Literature Synthesis (\Cref{sec:skill_literature})};
            \node (sk_hyp) [skill, below=0.7cm of hyp] {Hypothesis Generation (\Cref{sec:skill_hypothesis})};
            \node (sk_math) [skill, below=0.7cm of der] {Math/Symbolic Reasoning (\Cref{sec:math_skills_physics})};
            \node (sk_code) [skill, below=0.7cm of sim] {Coding \& Tool Use (\Cref{sec:code_skills_physics})};
            \node (sk_phys) [skill, below=0.7cm of ana] {Physical Consistency (\Cref{sec:skill_physical_ambiguity})};
            \node (sk_multi) [skill, below=0.7cm of com] {Multimodal Reasoning (\Cref{sec:advancing_multimodal})};
            \draw [skillarrow] (sk_lit.north) -- (lit.south);
            \draw [skillarrow] (sk_hyp.north) -- (hyp.south);
            \draw [skillarrow] (sk_math.north) -- (der.south);
            \draw [skillarrow] (sk_code.north) -- (sim.south);
            \draw [skillarrow] (sk_phys.north) -- (der.south);
            \draw [skillarrow] (sk_phys.north) -- (ana.south);
            \draw [skillarrow] (sk_phys.north west) -- (der.south east);
            \draw [skillarrow] (sk_phys.north east) -- (sim.south west);
            \draw [skillarrow] (sk_multi.north) -- (com.south);
            \draw [skillarrow] (sk_multi.north west) -- (ana.south east);
            \draw [skillarrow] (sk_multi.north west) -- (lit.south east);;
            \node (opp_lit) [opportunity, below=0.3cm of sk_lit] {Speed up survey};
            \node (opp_hyp) [opportunity, below=0.3cm of sk_hyp] {Novel connections};
            \node (opp_math) [opportunity, below=0.3cm of sk_math] {Error detection};
            \node (opp_code) [opportunity, below=0.3cm of sk_code] {Experiment control};
            \node (opp_phys) [opportunity, below=0.3cm of sk_phys] {Verify constraints};
            \node (opp_multi) [opportunity, below=0.3cm of sk_multi] {Diagram to code};
        \end{tikzpicture}
    }
    \caption{A schematic workflow of theoretical physics research (top row, blue), potential LLM capabilities (middle row, green), and key opportunities (bottom row, orange). Tool use capability connects with experimental research through automated instrument control and data analysis.
    }
    \label{fig:workflow_skills}
\end{figure*}
\subsection{Research Stages \& Skills}
\label{sec:workflow_example}
A typical workflow in physics research often involves several stages, as depicted in \Cref{fig:workflow_skills} (top row). These stages are generally iterative and involve collaboration among various researchers with different backgrounds and skill sets.
Scientific inquiry typically proceeds through an iterative workflow that begins with \textit{literature review and problem identification}, where existing work is surveyed to assess the state of the art and uncover open questions or inconsistencies. Based on this foundation, researchers engage in \textit{hypothesis formulation and model building}, proposing new ideas, constructing models to capture physical phenomena, and defining the assumptions that frame their scope. These models are then subjected to \textit{analytical derivation}, involving mathematical analysis, symbolic reasoning, and numerical calculations to extract predictions. Complementing this, \textit{simulation and computational experiments} are employed to test model behavior and guide the design of physical experiments, for instance simulating the BKT transition in a quantum XY model~\cite{Ding1990KosterlitzThouless} before performing quantum optical experiments. The resulting data undergo \textit{analysis and interpretation of results}, where findings are compared with prior work to generate physical insights. This process is inherently cyclical, requiring \textit{iteration} of the above stages until the problem is satisfactorily addressed. Finally, the outcomes are consolidated through \textit{communication}, including the preparation of papers and presentations to disseminate the results.

\subsection{Opportunities and Challenges for LLMs in Physics Research}
\label{sec:llm_opportunies}
The intersection of AI and physics is not new~\cite{Carleo2019Machine}, but the advent of powerful LLMs introduces the potential to help address persistent bottlenecks in physics research---especially for tasks demanding enormous time investment or the processing of vast information streams.
LLMs might assist physics research in at least two primary modes: (1) automating repetitive tasks such as literature review and well-defined calculations (see \Cref{sec:skill_literature,sec:code_skills_physics,sec:math_skills_physics}), and (2) sparking new ideas through human-AI collaboration, where AI agents might provide alternative perspectives (see \Cref{sec:skill_hypothesis,sec:good_ai_physicist}).

While LLMs have shown remarkable growth in theorem proving~\cite{Azerbayev2023Llemma,Trinh2024Solving,Romera-Paredes2024Mathematical} and augmenting biochemical research~\cite{M.Bran2024Augmenting,Sarwal2025Benchmark}, physics poses unique challenges. Unlike formal mathematics with its focus on axiom-based proof~\cite{Azerbayev2023Llemma}, theoretical physics centrally involves constructing models, making justified approximations (e.g., when to approximate $\sin\theta \approx \theta$ for small angles or apply perturbation theory where a Hamiltonian $H = H_0 + \lambda V$ is expanded in powers of a small parameter $\lambda$), seeking validation in experiments, and even designing new models from scratch with physical intuition. These represent the highly challenging task of connecting mathematical abstractions to physical reality~\cite{Hagendorff2024Machine,Schaeffer2023Are,Wu2024Reasoning}. Physics often involves mathematical problems that, while formally straightforward, gain complexity and nuance from their physical context, where mathematical rigor alone is insufficient.
For example, diagonalizing a $2 \times 2$ matrix is trivial in standard linear algebra. However, in topological condensed matter physics, such a matrix might represent the Bloch Hamiltonian of a Chern insulator, $H(\mathbf{k}) = d_x(\mathbf{k})\sigma_x + d_y(\mathbf{k})\sigma_y + d_z(\mathbf{k})\sigma_z$, where $\sigma_i$ are Pauli matrices and $\mathbf{d}(\mathbf{k})$ is a vector function of momentum $\mathbf{k}$. Finding eigenvalues $E_\pm(\mathbf{k}) = \pm |\mathbf{d}(\mathbf{k})|$ is straightforward mathematically, but the physics lies in how the winding of $\mathbf{d}(\mathbf{k})$ determines topological invariants like the Chern number~\cite{Thouless1982Quantized}, which dictates quantized Hall conductivity~\cite{Klitzing1980New}. LLM agents must learn to connect mathematical formulas with their underlying physical interpretation to contribute meaningfully.

\section{Skill Analysis for Physics Reasoning}
\label{sec:skill_analysis_physics}
Despite the emerging ecosystem of scientific reasoning benchmarks from general scientific knowledge~\cite{Hendrycks2021Measuring,Sun2023SciEval} to physics-specific reasoning~\cite{Chung2025Theoretical,Zhang2025PhysReason,Wang2024SciBench,seephys,wang2025cmphysbenchbenchmarkevaluatinglarge}, they focus primarily on exam-like problems with single definitive answer for easy verification. These benchmarks could measure test-taking capability but do not capture the full complexity of real-world physics research involving novel tasks such as deriving properties of fundamentally new physical models, modifying simulation code based on a new paper, and interacting with other experts to explore interdisciplinary open-ended problems. We need more benchmarks analogous to SWE-Bench~\cite{Jimenez2024SWEbench} on a full-cycle research workflow to gauge how LLMs perform in tackling open-ended research in a real-world scenario~\cite{Starace2025PaperBench,liu2025mimickingphysicistseyeavlmcentric,zheng2025newtonbenchbenchmarkinggeneralizablescientific,pan2026cmtbenchmark,wang2026cmphysbench}. Existing agentic benchmarks evaluate substeps of research---writing code snippets or solving well-defined problems ---but not the complete non-trivial reasoning required to produce frontier publication-level results, such as deriving a novel theorem or replicating the technical appendix of a research paper. Furthermore, developing benchmarks based on expertise from real-world researchers such as FrontierMath~\cite{Glazer2024FrontierMath} or Humanity's Last Exam~\cite{Phan2025Humanitys} within a carefully maintained ecosystem of domain experts~\cite{PathintegralInstitute2025Benchscience}, is key to probing the limits of AI reasoning for scientific discovery beyond solving close-ended Olympiad exam questions. Early attempts at research-level replication reveal a characteristic failure pattern: models correctly execute standard procedures (e.g., routine operator expansions) but fail at the key non-standard step requiring genuine insight---discovering a novel proof technique or recognizing a technically delicate prerequisite before the main argument can proceed.

In this section, we discuss concrete key skills needed for physics research with current limitations. We categorize these skills to better understand the multi-dimensional potential and challenges in modern physics research.

\subsection{Mathematical and Symbolic Reasoning}
\label{sec:math_skills_physics}
\textbf{Skill} Performing algebraic manipulation, calculus (differentiation, integration), linear algebra (matrix operations, tensor contractions like $T^{ijk}S_{jlm} = R^{ik}_{lm}$), and solving differential equations essential for theoretical physics.
\textbf{Analysis} Next-token prediction inherent to LLMs can lead to cascading errors in complex mathematical operations. Despite saturation on legacy benchmarks like MATH~\cite{Hendrycks2021Measuring}, errors are frequently observed in algebra and calculus~\cite{Davis2025Testing,Davis2024Testing}. They also struggle with unit consistency (e.g., mixing SI and natural units), thereby raising questions about their reliability for research-level derivations (e.g., evaluating path integrals $\int \mathcal{D}\phi e^{iS[\phi]/\hbar}$)~\cite{Pan2025Quantum}.

\subsection{Beyond Math: Physics-Specific Reasoning Skills}
\label{sec:physics_specific_skills}
We outline skills unique to understanding physical context, principles, and common practices, ordered roughly from currently more reliable to less reliable (or more complex) for LLMs.
\subsubsection{Conceptual Framework, Formula Retrieval, and Application}
\label{sec:skill_conceptual_formula}
\textbf{Skill} Articulating physics concepts, principles, and theories in natural language, adapting to specific notations; identifying and applying general physics formulas to well-structured problems.
\textbf{Analysis} LLMs can generate textbook-style explanations through summarization, yet this apparent understanding can be superficially derived from statistical correlations rather than causal models of physical laws~\cite{Hagendorff2024Machine,Schaeffer2023Are}. This appears in explanations with subtle inaccuracies or missed assumptions (e.g., in the context of perturbation theory, failing to state the conditions for its validity)~\cite{Zhang2023Sirens}. LLMs have shown promising progress in applying formulas to well-defined problems mirroring textbook examples~\cite{Qiu2025PHYBench,Lewkowycz2022Solving,Zhang2025PhysReason}, but they may resort to memorized solutions rather than reasoning from first principles when confronted with novel variants of the same problems. Recent work~\cite{Srivastava2022Imitation,Wu2024Reasoning,Huang2025MATHPerturb} shows that simple perturbations to problem statements can cause significant performance decay, revealing models' fragile understanding of systematic solution strategies. Their ability to choose appropriate approximations or understand the domain of validity for a given formula remains limited.
\subsubsection{Mathematical Deduction and Reasoning by Special Cases and Analogies}
\label{sec:skill_math_physics_context_special}
\textbf{Skill} Applying mathematical tools adaptively to respect the physical constraints and interpretations of variables and operations; simplifying complex problems by considering special or limiting cases, or by drawing analogies to simpler, well-understood physical systems.
\textbf{Analysis} This involves applying mathematical tools correctly while respecting the physical constraints and interpretations of variables and operations. For instance, correctly applying vector calculus to electromagnetic fields requires not just knowing the formulas for divergence or curl, but understanding what these operations mean for fields, sources, and boundaries in a physical system. LLMs are improving but can still falter in maintaining this contextual awareness through complex derivations.
A challenge here is the potential for LLMs to exhibit overcomplication bias. Furthermore, behavioral tuning (e.g., for verbosity or specific output formats like Markdown) might inadvertently reduce their core reasoning capabilities, an effect sometimes termed an ``alignment tax''~\cite{Askell2021General}. The default system prompts of general-purpose LLMs may also not elicit the concise, formal style of mathematical physics, potentially hiding their performance on complex derivations. For shorter calculations, some LLMs have struggled with tasks like counting the number of `r's in the word ``strawberry'' or computing `9.9-9.11'. In physics, there are many notations whose rules differ dramatically, and LLMs should understand the context and apply the correct rule, such as the normal ordering notation with $\colon\cdot\colon$ from quantum many-body physics.

A common strategy in physics research is to gain intuition about a complex problem by analyzing simpler, solvable special cases (e.g., 1D version of a 2D problem, specific symmetry points in parameter space) or by relating it to analogous systems (e.g., mapping a quantum spin system to a classical statistical mechanics model). LLMs show some ability to follow instructions to analyze special cases. For example, given a general expression for the magnetic susceptibility $\chi(T)$, an LLM might be able to evaluate its behavior as $T \to 0$ (e.g., Curie's law $\chi \propto 1/T$ for paramagnets~\cite{Kittel2004Introduction}) or $T \to \infty$. However, spontaneously identifying fruitful special cases or insightful analogies remains an underdeveloped skill.

\textbf{Example: Interacting Systems} Consider a complex interacting quantum system with Hamiltonian $H = H_{\text{kin}} + H_{\text{int}}$. A physicist might first analyze the noninteracting limit (setting $U=0$ in $H_{\text{int}}$) to build intuition. An LLM could be guided to do this, but proactively suggesting to consider the case where $U=0$ or recognizing analogies (e.g., to the Ising model) demonstrates higher-level scientific reasoning.
\subsubsection{Physical Consistency, Constraint Satisfaction, and Navigating Ambiguity}
\label{sec:skill_physical_ambiguity}
\textbf{Skill} Ensuring solutions respect fundamental physical principles (e.g., conservation laws like $dE/dt=0, d\mathbf{P}/dt=0$, dimensional consistency, causality, symmetries) and problem-specific constraints; recognizing ambiguity in problem statements or scientific texts, making justified assumptions to resolve ambiguity, or querying for clarification.
\textbf{Analysis} A critical aspect of physics reasoning is ensuring solutions are physically sensible. LLMs must learn to self-check outputs against fundamental physical laws (e.g., conservation of energy, momentum, charge) and problem-specific constraints (e.g., boundary conditions like $\psi(x=\pm L/2) = 0$ for a particle in a box~\cite{Griffiths2018Introduction}, symmetries of the Hamiltonian such as $[H, P]=0$ if parity $P$ is conserved). This includes ensuring dimensional consistency of equations (e.g., verifying that terms being added have the same physical units) and respecting fundamental symmetries. Developing this ``physical common sense'' is needed. Current LLMs may generate solutions that are mathematically plausible but physically violate such principles if not carefully guided or checked. Self-correction techniques~\cite{Madaan2023SelfRefine, Shinn2023Reflexion, Saunders2022SelfCritiquing} must be adapted to evaluate physical plausibility alongside logical consistency.

\textbf{Example: System-Bath Modeling.} In modeling system-bath interactions for quantum spin systems with Hamiltonian $H = H_S(\{\sigma_i\}) + H_B(\{\tau_j\}) + H_{SB}(\{\sigma_i\}, \{\tau_j\})$, an LLM might erroneously place system spins $\{\sigma_i\}$ and bath spins $\{\tau_j\}$ on the same lattice sites if not explicitly prohibited. This is physically nonsensical for typical models where system and bath are distinct subsystems. Such errors reveal current gaps in LLMs' physical intuition about subsystem independence.
Physics research often involves nuanced statements or different notations relying on implicit context. When faced with choices that lead to different solution paths, a human scientist typically seeks clarification, yet LLMs tend to randomly pick one path without justification. This extends to interpreting under-specified problems common in physics, akin to Fermi problems (order-of-magnitude estimations often based on ambiguous information)~\cite{Weinstein2010Fermi}, where making justified assumptions is essential.

\textbf{Example: Notational Ambiguity} A research note might define a spin Hamiltonian $H_s = -J \sum_{\langle i,j \rangle} \sigma_i^z \sigma_j^z$, then describe a Jordan-Wigner transformation to a fermionic Hamiltonian $H_f = -J \sum_{\langle i,j \rangle} (2c_i^\dagger c_i - 1)(2c_j^\dagger c_j - 1) + \dots$. For brevity, an author might informally refer to both as `$H$' in different contexts. LLM agents often confuse properties valid for $H_s$ (acting on spin Hilbert space) with those for $H_f$ (acting on Fock space). They might attempt to `correct' notation or apply operations valid for $H_s$ to the transformed $H_f$ if they fail to track the change in underlying variables and Hilbert space, leading to cascading errors.
\subsubsection{Making Justified Physical Approximations}
\label{sec:skill_approximations}
\textbf{Skill} Selecting appropriate levels of approximation based on physical context, stating assumptions explicitly, and understanding the domain of validity.
\textbf{Analysis} Exact solutions are rare; progress often hinges on making well-justified approximations. LLMs need to select appropriate approximation levels (e.g., classical vs. quantum, relativistic vs. nonrelativistic, perturbative expansions, mean-field theory). They may default to standard textbook approximations (like the ideal gas law $PV=nRT$~\cite{Reif1965Fundamentals} or the harmonic oscillator potential $V(x) = kx^2/2$) without critically evaluating their validity for the specific problem context or stating the conditions under which they hold. This includes complex expansions like those in stochastic calculus or advanced quantum field theory, where the choice of approximation scheme is nontrivial.

\textbf{Example: Perturbation Theory} Consider a quantum system with a Hamiltonian $H = H_0 + \lambda V$, where $H_0$ is exactly solvable (e.g., a free particle or harmonic oscillator), $\lambda$ is a small dimensionless perturbation parameter, and $V$ is the perturbation potential. An LLM might be asked for the first-order correction to the ground state energy $E_0^{(0)}$ of $H_0$. It should retrieve the standard formula from time-independent perturbation theory: $E_0^{(1)} = \lambda \langle \psi_0^{(0)} | V | \psi_0^{(0)} \rangle$ (see, e.g.,~\cite{Sakurai2017Modern}), where $\psi_0^{(0)}$ is the ground state eigenfunction of $H_0$. However, a crucial aspect is understanding the conditions for the validity of perturbation theory, such as $|\lambda \langle \psi_m^{(0)} | V | \psi_n^{(0)} \rangle| \ll |E_m^{(0)} - E_n^{(0)}|$ for $m \neq n$. An LLM might apply the formula without checking or stating this crucial assumption, or struggle to identify the appropriate $H_0$ and $V$ if the problem is not explicitly presented in this standard perturbative form (a Taylor expansion in $\lambda$).
\subsection{Developing Taste and Gracefulness}
\label{sec:good_ai_physicist}
\textbf{Skill} Exhibiting good research ``taste'', such as resorting to mathematically elegant explanations by Occam's Razor and avoiding unnecessary complexity.
\textbf{Analysis} While solving a problem is hard, solving it elegantly or finding the most insightful approach is much harder. A ``good'' physicist would not be satisfied with a brute-force answer but would strive for solutions that are simple, generalizable, and offer deeper understanding. This relates to developing a form of ``research taste''. Current LLMs may sometimes opt for overly complex or brute-force approaches if not guided. Training LLMs to recognize and prefer elegant or simpler solutions, perhaps through reinforcement learning from human feedback that rewards such qualities, could be an important direction~\cite{Bai2022Training}. Interpretability studies can also help understand how LLMs arrive at solutions and whether they are employing physical reasoning or relying on superficial pattern matching~\cite{Hanna2023How}.

\textbf{Example: Exploiting Symmetry}
Consider calculating the expectation value of position $\hat{x}$ for a particle in a symmetric one-dimensional potential $V(x) = V(-x)$, such as the harmonic oscillator $V(x) = m\omega^2 x^2/2$ or an infinite square well centered at origin. For an energy eigenstate $|\psi_n\rangle$, the wavefunction $\psi_n(x)$ has definite parity: either even ($\psi_n(-x) = \psi_n(x)$) or odd ($\psi_n(-x) = -\psi_n(x)$). Consequently, $|\psi_n(x)|^2$ is always even. The expectation value is $\langle \hat{x} \rangle_n = \int_{-\infty}^{\infty} x |\psi_n(x)|^2 dx$. Since $x$ is odd and $|\psi_n(x)|^2$ is even, their product is odd. The integral of an odd function over a symmetric interval is zero, thus $\langle \hat{x} \rangle_n = 0$ (a standard result discussed in, e.g.,~\cite{Shankar1994Principles}).
An LLM might attempt brute force: find explicit $\psi_n(x)$ (e.g., Hermite polynomials for the harmonic oscillator) and integrate, risking calculation errors. A `good AI physicist' would recognize the symmetry argument to immediately conclude $\langle \hat{x} \rangle_n = 0$ without detailed calculation. Training LLMs to identify and use such symmetries reflects deeper physical understanding and leads to more elegant problem-solving. This symmetry principle extends profoundly in physics~\cite{Gross1996Role}, from continuous symmetries to discrete ones (e.g., CP violation~\cite{Lee1956Question}).
\subsection{Code Generation and Execution for Physics}
\label{sec:code_skills_physics}
\textbf{Skill} Physics-aware code generation that correctly translates physical models and algorithms, bridging theory, computation, and experiment.
\textbf{Analysis}
LLMs can generate code (NumPy/SciPy) for Monte Carlo simulations in solid state physics and molecular dynamics, perform numerical analysis of equations~\cite{Cherian2024LLMPhy}, and assist with data analysis~\cite{Gao2023PAL}, helping to accelerate prototyping. They might assist in maintaining/extending legacy code (e.g., Fortran in large collaborations~\cite{Zhou2024Proofofconcept}) or translating to modern languages. This capability could help bridge theory, computation, and experiment: a theorist might use an LLM to quickly prototype a simulation for a new model; an experimentalist might use it to apply computational analysis to their data without extensive programming expertise.
However, physics-aware code generation~\cite{Tian2024SciCode} demands correct translation of physical models. For instance, implementing the Hubbard model $H = -t \sum_{\langle i,j \rangle, \sigma} (c_{i\sigma}^\dagger c_{j\sigma} + \text{h.c.}) + U \sum_i n_{i\uparrow} n_{i\downarrow}$~\cite{Auerbach1994Interacting} requires understanding its Hilbert space, symmetries (particle number, $S_z$ conservation), and numerical algorithms (exact diagonalization, Quantum Monte Carlo~\cite{Becca2017Quantum}). A naive LLM agent might miss crucial physical constraints like fermionic anticommutation rules $\{c_{i\sigma}, c_{j\sigma'}^\dagger\} = \delta_{ij}\delta_{\sigma\sigma'}$ or boundary conditions (e.g., periodic $c_{N+1} = c_1$). Similarly, translating Lattice Gauge Theory (LGT) formalisms~\cite{Wilson1974Confinement}, like the SU($N_c$) Hamiltonian $H = \frac{g^2}{2a} \sum_{l, \alpha} E_l^\alpha E_l^\alpha - \frac{1}{ag^2} \sum_p \text{ReTr}(U_p)$ (where $E_l^\alpha$ are electric field operators, $U_p$ plaquette operators, $a$ lattice spacing, $g$ coupling), into code requires handling complex group symmetry SU($N_c$).

\section{LLM Techniques as Augmentation for Physics Research}
\label{sec:llm_engineering_skills}
Various techniques in LLM reasoning can be adapted to tackle several common tasks within physics research, as we detail in this section.
\subsection{Literature Review by Long-Context  Retrieval}
\label{sec:skill_literature}
By leveraging Retrieval-Augmented Generation (RAG)~\cite{Lewis2020Retrievalaugmented}, frontier agentic research systems like DeepResearch~\cite{OpenAI2025Introducing} can access massive up-to-date literature. The rise of long-context LLMs~\cite{Team2024Gemini} enables workflows requiring comprehensive summarization across various data sources such as multiple \textit{Physical Review} papers, PhD theses, or graduate-level textbook chapters (e.g., following the derivation of the Bethe Ansatz solution~\cite{Bethe1931Zur,Essler2005Onedimensional} for the 1D Heisenberg model $H = J \sum_i \mathbf{S}_i \cdot \mathbf{S}_{i+1}$). However, practical limitations persist as performance often degrades as context length increases (the ``lost in the middle'' phenomenon~\cite{Liu2023Lost}), and models can be easily distracted by irrelevant information~\cite{Shi2023Large}. Effectively combining information beyond the context window remains an open challenge~\cite{Beltagy2020Longformer,Skarlinski2024Language} despite emerging techniques like compressing conversation history.\footnote{\url{https://www.anthropic.com/engineering/effective-context-engineering-for-ai-agents}}

\subsection{Exploratory Reasoning by In-Context Learning}
\label{sec:skill_few_shot}
LLMs can adapt their behavior based on in-context demonstrations~\cite{Brown2020Language,Madaan2022Language}. For example, LLM agents could infer how to tackle a particular type of equation from a few examples (e.g., the time-independent Schr{\"o}dinger equation $(-\hbar^2/2m \cdot \psi'' + V\psi) = E\psi$ for different potentials $V(x)$ like the harmonic oscillator $V(x) =m\omega^2x^2/2$) and then apply a similar methodology to a new potential, such as microwave shielding for cold molecules~\cite{Chen2023Fieldlinked,Deng2023Effective}, where experimental setups require analyzing a new long-range potential. Using few-shot examples of similar long-range potential analyses, an LLM could help researchers apply established analysis procedures to these novel experimental configurations.
This is particularly relevant in pursuit of new physics that involves new conditions or classes of models where the general solution methodology is known and the solutions are verifiable. Such adaptation is non-trivial: new settings require identifying physically reasonable assumptions (e.g., when generalizing from Markovian setting to Non-Markovian~\cite{Cubitt2015Stability})---not merely mathematically consistent ones---that capture the relevant physics, distinguishing theoretical physics from pure mathematics. LLM agents could study multiple variants of the same problem or multiple solution paths for the same conjecture simultaneously to help human researchers.
\subsection{Tool Use and Reliable Scientific Reasoning by Self-Reflection}
\label{sec:tool_integration_reliability}

\textbf{Tool Use} LLMs are not inherently calculators or symbolic reasoners, but they can effectively use external tools like symbolic math engines (Mathematica, SymPy), numerical libraries (via code execution), or databases via dedicated portals such as Model-Context Protocols (MCP)~\cite{Shen2023HuggingGPT,Patil2023Gorilla,Anthropic2024Introducing,PathintegralInstitute2025Mcpscience}. Models need to learn when and how to call these tools effectively, formulate valid queries for them (e.g., correctly translating a subproblem like ``calculate $\int_0^\infty x^2 e^{-ax} dx$ for $a>0$'' (textbook result~\cite{Arfken2013Mathematical}) into \verb|Integrate[x^2 Exp[-a x], {x, 0, Infinity}]| for Mathematica), and interpret their output correctly within the physics context.
Tool use allows for a more dynamic, nonsequential workflow: LLM agents can query a tool, analyze the output, and then decide on subsequent actions, effectively optimizing their solution path.
A productive pattern for symbolic physics is: use Mathematica to test ad-hoc linear algebra constructions, verify candidates against constraints, and iteratively converge toward a short, verifiable algorithm---then extract analytical understanding (e.g., recognizing the underlying abstract algebra) that generalizes beyond the specific instance~\cite{Cirac2021Matrix}.
A productive pattern for symbolic physics is: use Mathematica to test ad-hoc linear algebra constructions, verify candidates against constraints, and iteratively converge toward a short, verifiable algorithm---then extract analytical understanding (e.g., recognizing the underlying abstract algebra) that generalizes beyond the specific instance~\cite{Cirac2021Matrix}

\textbf{Self-Reflection} LLMs suffer from hallucinations or confabulations, which may produce factually incorrect information that sounds plausible at first~\cite{Zhang2023Sirens,Bottou2025Fiction}. This can lead to flawed conclusions or even potentially dangerous outcomes in an experimental setting. Ensuring the factual accuracy and logical consistency of LLM outputs, especially for complex reasoning chains, remains a major challenge~\cite{Lightman2023Lets}. Techniques like self-critique~\cite{Saunders2022SelfCritiquing} and RAG~\cite{Lewis2020Retrievalaugmented} with physics-specific knowledge bases show promise for improving factual accuracy but need further development for scientific domains.
Self-reflection~\cite{Madaan2023SelfRefine,Shinn2023Reflexion,Saunders2022SelfCritiquing} by external modules or human oversight~\cite{Lightman2023Lets,McAleese2024LLM,Kirchner2024ProverVerifier} has shown promising performance gains on scientific tasks~\cite{Lightman2023Lets,Wang2022SelfConsistency}. This is particularly valuable for catching logical inconsistencies, sign errors in derivations, or violations of conservation laws.

Multi-agent simulations~\cite{Zhang2025Collective,Du2023Improving,Wu2023AutoGen,Zou2025Agente,MultiAgent4Science} open the door for streamlining verification, where specialized agents verify different aspects of a derivation (e.g., one agent checks the validity of approximations, another verifies operator inequalities or bound arguments). A promising pattern is the deriver-critic loop, where one agent generates derivations while another critiques them for errors---either autonomously or with a human expert validating the critic's annotations. Crucially, critic agents should operate without access to intermediate generation artifacts to ensure independent verification. Generating multiple candidate derivations and using a grader agent to select the best can further improve reliability and may surface novel proof strategies. Such architectures might help accelerate the hypothesis-verification cycle of scientific discovery.

\section{Open Directions and Opportunities for LLM agents}
\label{sec:future_llms}
\subsection{Advancing Multimodal Reasoning}
\label{sec:advancing_multimodal}
Physics is inherently multimodal, relying on text, equations, diagrams, and various forms of data. LLM agents must evolve to efficiently integrate these diverse information types by parsing, interpreting, and generating specialized visual representations such as Feynman diagrams (see \Cref{fig:feynman_diagram_example}), tensor network notations~\cite{Cirac2021Matrix} (as shown in the example below), dual unitary circuit diagrams~\cite{Bertini2019Exact} (used in studies of quantum chaos), and phase diagrams.

Current vision-language models show potential in interpreting general plots but struggle with highly specialized physics notations~\cite{Bang2023Multitask}. The ability to seamlessly reason across modalities---for example, connecting a mathematical formalism with its graphical representation and experimental data---is important. This extends to translating diagrams into executable programs (e.g., a quantum circuit diagram into code for a quantum simulator) and assisting in graphical proofs or derivations~\cite{Jaffe2018Mathematical}. Recent advances like OpenAI-GPT5 demonstrate improved image analysis by calling tools to crop/zoom-in images, but understanding the deeper semantics of physics visualizations requires further progress and careful benchmarking.

\textbf{Example: Tensor Network Diagram}
Understanding and manipulating diagrams in specialized fields, such as tensor network notation used in quantum information and condensed matter theory, illustrates the type of complex visual-symbolic language that future LLMs should handle. Consider the following tensors (all indices are 3D, indexed from 0):

$\diagram{\draw (1,0) -- (0,0) -- (0,-1); \draw[ten] (0,0) circle (1/2); \draw[ten] (0,0) node {$A$}; \draw[ten] (1.3,0) node {$\scriptstyle i$}; \draw[ten] (0,-1.3) node {$\scriptstyle j$};} = i^2-5j$,
$\diagram{\draw (0,0) --(0,1) (0,0) -- (.75,.75) (0,0) -- (1,0); \draw[ten] (0,0) circle (1/2); \draw[ten] (0,0) node {$B$}; \draw[ten] (0,1.3) node {$\scriptstyle i$}; \draw[ten] (1,1) node {$\scriptstyle j$}; \draw[ten] (1.3,0) node {$\scriptstyle k$};} = 4^{ij}-k$,
$\diagram{\draw (-1,0) -- (0,0) -- (0,1); \draw[ten] (0,0) circle (1/2); \draw[ten] (0,0) node {$D$}; \draw[ten] (-1.3,0) node {$\scriptstyle i$}; \draw[ten] (0,1.3) node {$\scriptstyle j$};} = j$,
$\diagram{\draw (0,0) --(-1,0) (0,0) -- (-.75,-.75) (0,0) -- (0,-1); \draw[ten] (0,0) circle (1/2); \draw[ten] (0,0) node {$C$}; \draw[ten] (-1.3,0) node {$\scriptstyle i$}; \draw[ten] (-1,-1) node {$\scriptstyle j$}; \draw[ten] (0,-1.3) node {$\scriptstyle k$};} = ijk$,
$\diagram{\draw (0,0)--(1.5,1.5)--(1.5,0)--(0,0)--(0,1.5)--(1.5,1.5); \draw[ten] (0,0) circle (1/2); \draw[ten] (1.5,0) circle (1/2); \draw[ten] (0,1.5) circle (1/2); \draw[ten] (1.5,1.5) circle (1/2); \draw (0,1.5) node {$A$}; \draw (0,0) node {$B$}; \draw (1.5,0) node {$D$}; \draw (1.5,1.5) node {$C$};}$.

The task is to calculate the value of the last tensor network (perturbed from~\cite{Bridgeman2017Handwaving}). An ideal LLM assistant would parse the diagram, identify tensors and connectivity, translate to an algebraic expression $\sum_{a,b,c,d,e,f} A_{af} B_{abc} D_{cd} C_{fed}$, substitute definitions, and compute via generated code. LLMs could assist by creating such diagrams from text or formula descriptions.
Mastery of such visual-symbolic languages could extend to interpreting Feynman diagrams, parsing quantum circuit diagrams to determine their unitary evolution~\cite{Kos2021Correlations}, or graphical proofs~\cite{Jaffe2018Mathematical}.

\subsection{Developing Agentic Capabilities for Scientific Discovery}
\label{sec:agentic_capabilities}
Future LLM systems may evolve into more autonomous agents performing full-cycle scientific tasks with greater independence under supervision. This requires long-running agents capable of sustained research over days or weeks, equipped with memory systems that enable continual learning---accumulating domain knowledge, tracking failed approaches, and building on prior derivations rather than starting from scratch with each query. Through human-in-the-loop feedback, such agents could also learn the reasoning style of the domain and the individual scientist, becoming more aligned and producing outputs that are easier to verify.
\subsubsection{Agentic AI for Hypothesis Generation and Verification}
\label{sec:skill_hypothesis}
Future AI agents might propose new models by analyzing anomalies~\cite{Wang2023Scientific} and inconsistencies among different theories, exploring multiple branches of a solution tree and alternative physical models or mathematical ans{\"a}tze, and then systematically validating each option.
By scanning through parameter spaces~\cite{Wu2023AutoGen, Zhou2023Language, Besta2024Graph, Liu2024Agentbench} guided by physical principles, AI-assisted scientific discovery~\cite{Lu2024AI, Wang2023Scientific} may eventually contribute to scientific hypothesis generation akin to AlphaGo Move 37~\cite{Silver2016Mastering,Silver2017Mastering}, though this remains an ambitious prospect.

A promising class of problems for AI agents are those where the search space is combinatorially large but candidate solutions can be verified in polynomial time. Physics offers many such problems: discovering quantum error-correcting codes satisfying Knill-Laflamme conditions~\cite{Knill1997Theory}, finding Hamiltonians with exact symmetries or dualities, classifying symmetry-protected topological phases~\cite{Chen2012SymmetryProtected}, or identifying integrable structures. In these settings, agents can enumerate candidates at scale while humans or automated checkers verify correctness against mathematical constraints---enabling certificate-backed discovery where every claimed result is independently validated.

Pre-LLM machine learning has demonstrated success in optimizing toward specific targets---minimizing quantum circuit depth for experimental feasibility or maximizing fidelity under noise constraints. What LLM agents add is the capacity for \emph{analytical reasoning}: beyond finding individual numerical solutions, they can identify structural patterns across verified instances and generalize them into infinite families---for example, deriving closed-form constructions of quantum error-correcting codes for arbitrary $n$ from a catalog of small-$n$ examples. Similarly, for variational methods where one seeks a wavefunction ansatz $|\Psi(\{\alpha_i\})\rangle$ that captures the physics (correlations, symmetries) while being verifiable by energy minimization, LLMs might suggest functional forms based on Hamiltonian structure---Matrix Product States for 1D systems~\cite{Cirac2021Matrix}, symmetry-adapted ans\"atze for lattice gauge theory~\cite{Zohar2018Combining,Bender2023Variational}, or non-Gaussian states~\cite{Shi2017Variational} requiring long analytical derivations. This transition from instance-level optimization to analytical generalization represents a qualitative shift in what AI can contribute to theoretical physics.

\subsubsection{Automated Simulation and Verifying Theoretical Results at Scale}
\label{sec:automated_design_future}
LLM agents might assist in optimizing experimental designs or simulation parameters, particularly where theoretical models can guide the process, to maximize information gain or test specific hypotheses~\cite{Kaiser2025Large}. This could involve suggesting appropriate measurement techniques (e.g., choosing between different spectroscopic methods to probe a material's electronic structure), identifying key parameters to calibrate in an experiment with a quantum gas microscope~\cite{Gross2017Quantum}, or even interfacing with automated cloud labs~\cite{Lu2024AI}.
A significant challenge would be for an LLM to assist in verifying highly complex proofs in mathematical physics, such as Hastings's proof of the super-additivity of Holevo information~\cite{Hastings2009Superadditivity}.
Such verification is particularly important given that very technical results typically take years to fact-check, and errors in published proofs are not uncommon~\cite{Vidick2020It}.
\subsection{Fine-tuned LLM Physicists and Towards AI Physicists}
Specialized LLMs fine-tuned for physics could offer advantages over general-purpose models~\cite{Beltagy2019SciBERT}, prioritizing domain knowledge (e.g., quantum mechanics principles), eliminating irrelevant information (e.g., historical facts unrelated to physics), and focusing on physical reasoning patterns (e.g., dimensional analysis, order-of-magnitude estimation, symmetry arguments). Fine-tuning could involve supervised~\cite{Ouyang2022Training} and reinforcement learning~\cite{Trung2024ReFT}.
To train physics-dedicated LLMs, we need more nuanced and effective reward signals. Beyond simple pass/fail on benchmark problems, rewards should capture the quality of the reasoning process~\cite{Lightman2023Lets}, the physical insightfulness of solutions, and alignment with established scientific methodology by incorporating feedback from domain experts.

A long-term vision would be for LLM agents to become effective AI collaborators, or building modules of automated ``AI physicists'', capable of full-cycle capabilities including proposing novel research ideas, theorizing experimental phenomena, verifying hypotheses and assisting in all research stages~\cite{Wang2023Scientific,Lu2024AI}. Apart from significant technical advances, this would require a synergistic partnership where AI models augment human intellect, supported by suitable UI/UX and appropriate guardrails.
Effective collaboration also requires that agents communicate results in ways humans can meaningfully engage with. Supervisors have limited time; if agents merely produce piles of derivations without conveying the discovery process and key insights, humans are reduced to mechanical checking rather than scientific understanding. With current success rates, where many agent outputs contain errors, humans would be even more discouraged from investing effort to parse them. Agents should mirror how theoretical physics groups operate: presenting results at varying levels of detail, highlighting why certain approaches work, and enabling supervisors to probe assumptions---so that discoveries can be understood, critiqued, and built upon.
Imagine investigating a novel material: an AI assistant might synthesize literature, formulate computational models using solid-state physics principles, generate simulation code, explore analytical approximations, and report results.
For theoretical physics, a distant prospect is to deploy AI agents for tackling open problems in the field~\cite{InstituteForQuantumOpticsAndQuantumInformationVienna2025Open,Horodecki2020Five}.
\section{Conclusion}
\label{sec:conclusion}
LLMs bear the potential to contribute meaningfully to modern physics research. Their potential to help accelerate scientific discovery, automate repetitive tasks, and assist in conceptual breakthroughs is considerable. However, realizing this potential requires substantial effort to address current limitations in rigorous reasoning, physical grounding, reliability, and multimodal understanding.
By fostering collaboration between physics and AI communities to develop specialized models, robust verification techniques, and effective human-AI interfaces, we can work toward using LLMs to contribute to expanding our understanding of the physical universe. Furthermore, applying LLMs to physics serves as a demanding testbed for studying LLMs, including interpretability~\cite{Hanna2023How},
faithfulness of reasoning, adversarial robustness, and scalable oversight~\cite{Bowman2022Measuring} for safety~\cite{Bengio2024Managing}.

%% file: Sections/appendix.tex
\appendix
\section{Risks, Limitations and Ethical Considerations}
\label{sec:risks_challenges}
\subsection*{Risk}
Over-reliance on LLMs without rigorous verification could embed subtle errors into research~\cite{Zhang2023Sirens}. The potential for LLMs to ``cheat'' reward functions during fine-tuning, producing plausible but physically invalid outputs (e.g., a simulation appearing to conserve energy due to numerical artifacts), requires careful alignment and robust evaluation~\cite{Bengio2024Managing,Anwar2024Foundational}.
Furthermore, depending too heavily on LLMs for tasks like mathematical derivations (e.g., routinely asking an LLM to compute integrals like $\int d^4k / (k^2-m^2+i\epsilon)^2$ instead of learning contour integration techniques), programming, or data interpretation could risk degrading these essential skills among physicists, especially those in training~\cite{Bender2020Climbing}. Deep intuition often arises from performing detailed calculations firsthand.
The history of scientific computing shows both warnings and reassurances: tools like Mathematica initially raised de-skilling concerns but ultimately enabled mathematicians to focus on higher-level work by automating repetitive calculations. Similarly, LLMs could potentially elevate physics research by handling routine tasks while humans focus on deeper insights---if used as augmentation rather than replacement for fundamental understanding.

\subsection*{Limitations}
\label{sec:limitations}
This position paper presents a high-level overview. The field of LLMs is rapidly evolving, and specific capabilities or limitations discussed may change quickly.
Due to the rapid evolution of LLMs, specific examples quickly become outdated. The selected examples are illustrative of general trends observed circa late 2024 and early 2025. The scope is necessarily limited to selected aspects of physics research, and specific examples may not generalize to all subfields.
\subsection*{Ethical Considerations}
\label{sec:ethics}
Generating plausible but incorrect claims requires rigorous validation. Training bias could steer research suboptimally. Responsible deployment and human oversight are required. Access and equity issues must be addressed to ensure broad availability of these tools across the global physics community.

\section{Related Works}
\paragraph{Non-Language Models Already Help Physics}
Machine learning (ML) is not new to physics~\cite{Carleo2019Machine}. Current applications include analyzing large experimental datasets (e.g., particle identification at the Large Hadron Collider~\cite{Guest2018Deep}), solving computational physics problems (e.g., finding ground states of quantum Hamiltonians like $H\psi = E\psi$~\cite{Carleo2017Solving,Glasser2018NeuralNetwork}), accelerating partial differential equation solvers~\cite{Karniadakis2021Physicsinformed}, and optimizing experimental controls (e.g., plasma shaping in fusion reactors~\cite{Degrave2022Magnetic}).
These applications typically involve supervised learning (classification, regression), unsupervised learning (clustering, dimensionality reduction, generative modeling), or reinforcement learning for specific, well-defined tasks. While powerful for specific tasks, these methods often differ from the requirements of open-ended theoretical exploration, complex multistep problem solving, or nuanced experimental design where LLMs might offer complementary advantages through their natural language interface and broad knowledge encoding~\cite{Schaeffer2023Are,Lewkowycz2022Solving}.
\section{More Examples}
\paragraph{Example: Notational Nuances}
Another example is the Bogoliubov-de Gennes (BdG) Hamiltonian, which often includes a $1/2$ prefactor by convention; LLMs might add or omit this factor inconsistently if not carefully prompted, thereby impacting all subsequent calculations even though the authors intend a different factor.
\paragraph{Example: Lattice Gauge Theory}
For example, implementing the pure gauge SU(2) Hamiltonian:
\begin{equation}
    H = \frac{g^2}{2} \sum_l \hat{E}^a_l \hat{E}^a_l + \frac{1}{2g^2} \sum_p \left(2 - \text{Tr}(\hat{U}_p + \hat{U}_p^\dagger)\right)
\end{equation}
where $\hat{E}^a_l$ are electric field operators on links $l$, $\hat{U}_p$ are plaquette operators, and $g$ is the coupling constant, requires translating abstract gauge theory concepts into concrete numerical algorithms that preserve gauge invariance and other symmetries.
A critical physical constraint in such simulations is ensuring that states satisfy Gauss's law, which in the quantum context becomes $\sum_{l \in \text{star}(n)} \hat{E}^a_l |\psi\rangle_{\text{phys}} = 0$ for each lattice site $n$ and each gauge group generator $a$. This constraint must be explicitly enforced in the code, typically by projecting onto the physical subspace or by adding an energy penalty term.

\paragraph{Example: Explaining the derivations}
A research paper may state a key result derived from an effective action, $S_{\text{eff}}[\phi_c]$, obtained by ``integrating out'' high-momentum modes $\phi_h$ from a full action $S[\phi_c, \phi_h]$. An LLM assisting a researcher could be tasked to elaborate on the formal path integral definition $e^{-S_{\text{eff}}[\phi_c]/\hbar} = \int \mathcal{D}\phi_h e^{-S[\phi_c, \phi_h]/\hbar}$. This elaboration might involve expanding the derivations with common evaluation techniques like saddle-point approximations or perturbative expansions of $S[\phi_c, \phi_h]$ around a background field, all while strictly adhering to the paper's specific notation for the classical fields $\phi_c$ and quantum fluctuations $\phi_h$.

\paragraph{Example: Diagonalizing a 2x2 Hermitian matrix}
When asked to diagonalize a general 2x2 Hermitian matrix $H = \begin{pmatrix} a & b-ic \\ b+ic & d \end{pmatrix}$ (where $a,b,c,d$ are real but have complex expressions), an LLM might default to a brute-force symbolic expansion of the characteristic determinant $\det(H - \lambda I) = 0$ to find eigenvalues, followed by solving systems of linear equations for eigenvectors. A `good AI physicist', however, might recognize the structure and suggest decomposing the matrix in the Pauli basis: $H = a_0 I + \mathbf{a} \cdot \boldsymbol{\sigma}$, where $I$ is the identity matrix, $\boldsymbol{\sigma} = (\sigma_x, \sigma_y, \sigma_z)$ are the Pauli matrices, $a_0 = (a+d)/2$, and $\mathbf{a} = (b, c, (a-d)/2)$. From this decomposition, eigenvalues ($a_0 \pm |\mathbf{a}|$) and eigenvectors (related to the direction of $\mathbf{a}$) can be read off with greater physical insight (e.g., connecting to spin precession in a magnetic field) and often less computation. Training LLMs to prefer such insightful decompositions over brute-force methods is key to developing AI assistants that contribute to more elegant theory and can reduce complex algebraic manipulations where errors might occur.

\paragraph{Example: Feynman diagram}
Interpreting a Feynman diagram~\cite{Feynman1949Theory} for Compton scattering ($\gamma e^- \to \gamma e^-$) (see, e.g.,~\cite{Peskin1995Introduction}) requires identifying incoming/outgoing photon (wavy lines) and electron (solid lines) lines, internal propagators (e.g., electron propagator $S_F(p) = i(\gamma \cdot p + m_e)/(p^2-m_e^2+i\epsilon)$, where $m_e$ is electron mass, $e$ is elementary charge, $\gamma^\mu$ are Dirac gamma matrices), and vertices (e.g., QED vertex factor $-ie\gamma^\mu$). An LLM should connect these diagrammatic elements to the mathematical terms in the scattering amplitude calculation according to Feynman rules.
\begin{figure}[h!]
    \centering
    \includegraphics[width=0.45\textwidth]{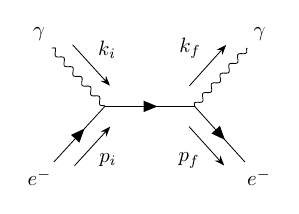}
    \caption{A Feynman diagram for Compton scattering ($\gamma e^- \to \gamma e^-$) in s-channel. LLMs should connect graphical elements (straight lines for fermions, wavy lines for bosons, and vertices for interactions) to mathematical terms in scattering amplitude calculations (e.g., propagators, vertex factors, external leg factors).}
    \label{fig:feynman_diagram_example}
\end{figure}

\begin{figure}[h]
    \centering
    \includegraphics[width=0.5\textwidth]{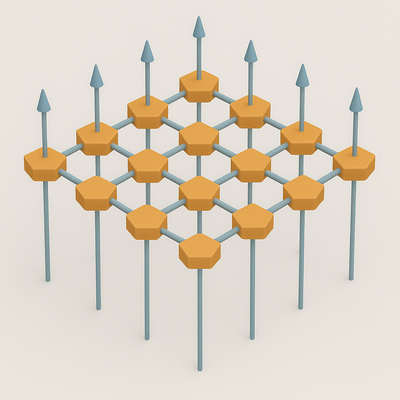}
    \caption{GPT-4o-generated PEPS network with incorrect or at least unconventional tensor connectivity. Proper PEPS tensors need 5 indices (4 virtual, 1 physical); many nodes lack required connections.}
    \label{fig:peps_ai_generated}
\end{figure}

\paragraph{Example: Physical Inaccuracy in AI-Generated Images}
Current AI image generators lack physical understanding, producing visually appealing but scientifically incorrect visualizations. Figure~\ref{fig:peps_ai_generated} shows GPT-4o's response to ``generate an image of a 3D modeling of a two dimensional projected entangled pair state tensor network (4 by 4 square lattice)''. The image violates PEPS structure: bulk tensors require exactly five indices (four virtual bonds to neighbors, one physical index), yet many nodes show incorrect connectivity. AI systems fail to encode the physical constraints---here, tensor network geometry and index structure.

\paragraph{Example: AI Material Physicist}
Imagine a physicist investigating a novel topological material. An `AI Physicist' assistant might: (1) Synthesize recent literature on related materials and their Berry curvature $\Omega_{n,xy}(\mathbf{k})$ calculations. (2) Assist in formulating a tight-binding Hamiltonian $H(\mathbf{k})$ for the new material based on its crystal structure (e.g., honeycomb lattice for graphene-like systems). (3) Generate Python code using libraries like Kwant or TightBindingTools.jl to numerically calculate the band structure $E_n(\mathbf{k})$ and Chern numbers $C_n = \frac{1}{2\pi} \int_{BZ} d^2k \Omega_{n,xy}(\mathbf{k})$. (4) If numerical results show unexpected edge states, it might help consider analytical approximations (e.g., a low-energy effective Dirac Hamiltonian $H_{eff} = v_F (k_x \sigma_y - k_y \sigma_x) + m \sigma_z$, where $v_F$ is the Fermi velocity and $m$ is a mass/gap parameter) to understand their origin. (5) Finally, it might create a slide deck summarizing these findings, including generating plots. This collaborative workflow, with the AI handling complex but definable subtasks under human strategic guidance, shows the potential~\cite{Wang2023Scientific}.

\paragraph{Example: Verifying analytical calculations by Mathematica}
Consider the Jordan-Wigner transformation, useful for 1D quantum spin systems. The transverse field Ising model Hamiltonian is $H = -J \sum_{\langle i,j \rangle} \sigma_i^z \sigma_j^z - h \sum_i \sigma_i^x$. The transformation maps spin operators $\sigma_i^\alpha$ to fermionic operators $c_j, c_j^\dagger$, e.g., $\sigma_j^z = 2c_j^\dagger c_j - 1$ and $\sigma_j^x = (\prod_{k<j} (1-2c_k^\dagger c_k)) (c_j + c_j^\dagger)$. An LLM might attempt this transformation and could be asked to verify parts via a Mathematica MCP, such as the anticommutation relation of the fermionic operators after the transformation $\{c_j, c_j^\dagger\}$. Using symbolic tools such as Mathematica, the probability of correctness can be increased, if LLMs become better at generating the correct query for a tool and interpreting its output for such (and more nontrivial) operator algebra.

\section{More Analysis}

\subsection{Building Better UI/UX for Human-centered AI}
\label{sec:building_better_ui_ux}
For LLMs to be effective collaborators, intuitive and efficient user interfaces (UIs) and user experiences (UXs) are essential for supervision, tracing, and trust-building. These interfaces should allow physicists to interact with LLMs naturally without extensive prompt engineering and should integrate with existing research workflows and tools (e.g., LaTeX editors, data analysis and simulation environments).
Future LLMs should also be robust to specific prompts and offer finer-grained controllability over their reasoning style, level of detail, and assumptions made. For instance, prompting an LLM to ``solve the Schr{\"o}dinger equation for a particle in a box'' might yield different solution forms depending on subtle phrasing. Ideally, an LLM should recognize standard conventions (e.g., specific boundary conditions $\psi(0)=\psi(L)=0$ for a box of length $L$) or prompt for these if ambiguous. Controllability would allow a physicist to specify, for example, ``provide a solution using separation of variables and show all steps for $H\psi = E\psi$ where $H = -\hbar^2/2m \cdot d^2/dx^2 + V(x)$ and $V(x)=0$ for $0<x<L$, $\infty$ otherwise'' versus ``give the energy eigenvalues $E_n = \frac{n^2\pi^2\hbar^2}{2mL^2}$ and normalized wavefunctions $\psi_n(x) = \sqrt{2/L}\sin(n\pi x/L)$ directly''. Such controllability is vital for making LLMs reliable and adaptive research assistants.
Effective UI/UX must go beyond simple chat interfaces. Physicists often work with extensive comments, annotations, and margin notes; interfaces supporting these natural workflows would be more effective. For supervising agents, UIs need robust mechanisms for managing experimental/simulation results, tracking context across long interactions (context management~\cite{Sumers2023Cognitive}), and accommodating human-in-the-loop intervention. Given that LLM outputs can be verbose, tools for generating structured summaries with highlighting are needed. Integration with collaborative platforms (e.g., Overleaf-like features with LLM assistance for consistency checking in \LaTeX{} documents, or GitHub-style review tools for coding and derivations) would also be convenient.

\subsection{Running Physical Experiments}
\label{sec:future:inter_ctrl_exp}
While our primary focus is on theoretical physics, we acknowledge that in the long term, AI systems might also actively control experimental instrumentation, interpret sensory data in real-time, and adjust experimental parameters accordingly. This would require the seamless integration of perception (e.g., using computer vision to optimize laser beam path setup for quantum optics experiments where alignment precision is critical for data quality), reasoning (understanding the experimental progress and deciding which measurements to perform next based on acquired data), and action (subsequently adjusting the lensing setup with high-precision robotic arms) in the physical world. This long-term vision enables a dynamic integration of reasoning models with robotics and control theory, bridging high-level human-defined agenda with corresponding physical actions as envisioned by rising interest in Large Action Models (LAM)~\cite{Wang2025Large} and Embodied Intelligence~\cite{Liu2025Embodied}.